\documentclass[twocolumn,10pt]{article}

\usepackage{epsfig}
\usepackage{subfig}
\usepackage{booktabs}
\usepackage{alltt}
\usepackage{textgreek}

\title{\bf Temporal expression normalisation\\in natural language texts}
\date{} 
\author{Michele Filannino\\School of Computer Science\\The University of Manchester, UK\\\texttt{filannim@cs.man.ac.uk}}

\begin{document}

\maketitle

\begin{abstract}

Automatic annotation of temporal expressions is a research challenge of great interest in the field of information extraction. In this report, I describe a novel rule-based architecture, built on top of a pre-existing system, which is able to normalise temporal expressions detected in English texts. Gold standard temporally-annotated resources are limited in size and this makes research difficult. The proposed system outperforms the state-of-the-art systems with respect to TempEval-2 Shared Task (\textsc{value} attribute) and achieves substantially better results with respect to the pre-existing system on top of which it has been developed. I will also introduce a new free corpus consisting of 2822 unique annotated temporal expressions. Both the corpus and the system are freely available on-line\footnote{http://www.cs.man.ac.uk/\~{}filannim/}.

\end{abstract}



\section{Introduction}

In many domains, the possibility of using and interpreting temporal aspects and events is important in order to organise information. Temporal knowledge allows people to filter information and even infer temporal flows of events. Furthermore, it permits an improving of intelligence for question answering, information retrieval and information filtering systems.

A temporal expression \cite{Ferro01}, also called \emph{timex}, refers to every natural language phrase that denotes a temporal entity such as an interval or an instant. For example, in a sentence like \emph{``Italian prime minister Mario Monti said yesterday that the reform has been very successful.''} the phrase \emph{``yesterday"} is actually a temporal expression. Timexes elicit a binding between the natural language domain and the time domain because it is always possible to represent such expressions as a time point, interval or set using ISO 8601 standard\footnote{http://www.w3.org/TR/NOTE-datetime}. Temporal expressions could be of three different types \cite{Ahn05}: \emph{fully-qualified}, \emph{deictic} and \emph{anaphoric}.
\begin{description}

\item[Fully-qualified] A temporal expression is fully-qualified with respect to the binding when all the information required to infer a point in the time domain are fully included inside the expression. In this category the following expressions falls: {\em March 15 2001}, {\em 21st July 1985} or {\em 31/04/2011}. Fully-qualified expressions are the easiest to detect because of their rigid lexical form.

\item[Deictic] In this case, inferring the binding with the time domain necessarily requires to take into account the time of utterance (when the document has been written or when the speech has been given). Deictic expressions could not be properly associated to a precise time without that information. Typical deictic temporal expressions are: {\em today}, {\em yesterday}, {\em last Sunday} and {\em two months ago}.

\item[Anaphoric] These expressions can be mapped to a precise point in the time domain only taking into account temporal expressions previously mentioned in the text or during the speech. Examples of this category are: {\em March 15}, {\em the next week}, {\em Saturday}. The only difference between deictic and anaphoric expressions is the location of the temporal reference: for deictic expressions it is the time of utterance or publication, for anaphoric expressions it is a time previously evoked in the text or speech. Anaphoric expressions constitute a future challenge for the scientific research in this field.

\end{description}

For the sake of completeness, another kind of categorisation \cite{Poveda09} is also adopted in the field. It identifies the possible shapes of timexes with respect to their semantics and admits the following types: {\em time or date references}, {\em time references that anchor on another time}, {\em durations}, {\em recurring times}, {\em context-dependent times}, {\em vague references} and {\em times indicated by an event}.

In my taster project I focussed on the normalisation of fully-qualified and deictic temporal expressions. I will not use the last categorisation because of the fuzziness of boundaries among types.

\section{Background}

The idea of annotating temporal expressions automatically from texts appeared for the first time in 1998 \cite{Becher98}. This topic aroused an increasing interest with the proposal of a proper temporal annotation scheme \cite{Mani00}. The original aim was to make the annotation phase easier with respect to the previous scheme in order to collect annotated data and use the temporal information to enhance performances of question answering systems \cite{Mani00}. All the most recent systems \cite{Verhagen07, Verhagen10} proposed for the temporal expressions extraction task go through two different steps: identification and normalisation. This dichotomy has become universally accepted by the research community because it makes the extraction phase easier to approach \cite{Ahn05}.

In the identification phase the effort is concentrated on how to detect properly the sub-expressions that are real temporal expressions in natural language texts. This step is usually done by using machine learning techniques. Ahn et al. \cite{Ahn05} firstly used Conditional Random Fields \cite{Lafferty01} showing better performances with respect to a previous work \cite{Ahn07} in which they used Support Vector Machines \cite{Boser92}. Poveda et al. \cite{Poveda09} introduced a sophisticated Bootstrapping technique enhancing the recognition of temporal expressions while Mani et al. \cite{Mani00} used rules learned by a decision tree classifier, C4.5 \cite{Quinlan93}, and Ling and Weld \cite{Ling10} tried Markov Logic Network \cite{Domingos04}.

The second step is the normalisation. In this phase the main goal is to interpret the expression, extract the temporal information and represent it in a proper pre-defined format. The universally accepted standard for temporal expressions annotation is TimeML \cite{Pustejovsky03}. It provides a specification for the representation of temporal expressions and also events. In this work the normalisation task aims at producing the proper TimeML code that correctly represents the temporal information (see Figure \ref{TimeML_example}). This step is usually accomplished using rule-based approaches. Grover et al. \cite{Grover10} used a rule-based approach on top of a pre-existing information extraction system, whereas Str\"{o}tgen and Gertz \cite{Strotgen10} produces a small set of hand-crafted rules with an ad-hoc selection algorithm. UzZaman and Allen \cite{UzZaman10} produced a rule-based normaliser focussing just on \textsc{type} and \textsc{value} attributes of TIMEX3 tag (the one used to represent timexes).

\begin{figure*}
\begin{alltt}
<?xml version="1.0" ?>
<TimeML xmlns:xsi="http://www.w3.org/2001/XMLSchema-instance"
        xsi:noNamespaceSchemaLocation="http://timeml.org/timeMLdocs/TimeML_1.2.1.xsd">
    <DOCID>Example_document</DOCID>
    <DCT>2012, Manchester, Apr 17, 2012</DCT>
    <TITLE>Example document</TITLE>
    <TEXT>
        Italian prime minister Mario Monti
        <EVENT eid="e1" class="OCCURRENCE">said</EVENT>
        <MAKEINSTANCE eiid="ei1" eventID="e1" pos="VERB" tense="PAST" aspect="NONE" />
        <{\bf{TIMEX3}} tid="t1"{\bf type}="DATE"{\bf value}="2012-04-16">yesterday<{\bf{/TIMEX3}}>
        that the reform has been very successful.
        <TLINK eventInstanceID="ei1" relatedToTime="t1" relType="DURING" />
    </TEXT>
</TimeML>
\end{alltt}
\caption{Example of TimeML code. In the sentence there is a deictic temporal expression; ``yesterday" can be correctly annotated only taking into account the document creation time (DCT).}
\label{TimeML_example}
\end{figure*}

\section{Method}

The contribution of my short taster project is twofold. Firstly, I will illustrate a temporal expression corpus explicitly designed for the normalisation phase. Then I will describe the software architecture of a new normaliser built on top of a pre-existing one.

\subsection{Temporal expressions corpus}

Gold-standard temporally-annotated resources are very limited in general domain \cite{Derczynski12}, and even less in specific ones like medical, clinical and biological \cite{Galescu12}. In the last decade, different sources of annotated temporal expressions have been developed. Because of the rapid evolution of this research field, usually the sources differ even with respect to the annotation guidelines. This leads to the existence of different corpora not entirely compatible to each other.

The main difference among them consists in the tag used to annotate temporal expressions: TIMEX2 against TIMEX3. These two tags reflect totally different way of annotating the same temporal expressions leading to the impossibility of using both corpora at the same time.

I created a corpus of temporal expressions collecting all TIMEX3 tags in four different corpora: AQUAINT\footnote{http://www.ldc.upenn.edu/Catalog/docs/LDC2002T31/}, TimeBank 1.2\footnote{http://www.timexportal.info/corpora-timebank12}, WikiWars\footnote{http://www.timexportal.info/wikiwars} and TRIOS TimeBank v0.1\footnote{http://www.cs.rochester.edu/u/naushad/trios-timebank-corpus}. I extracted from each document all the possible temporal expressions and for each one I also saved the related document creation time, the type (\emph{DATE}, \emph{TIME}, \emph{SET} or \emph{DURATION}) and the normalisation provided by the human annotators. Then I compacted the corpus removing possible duplicates. With the expression \emph{duplicates} I refer to completely identical tuples, i.e. same text, same normalisation, same utterance time and same type.

I obtained a corpus of 2822 unique annotated temporal expressions. The Table \ref{excerpt_corpus} shows an excerpt of the corpus. Further information about the distribution of temporal expression types in it is provided in Table \ref{corpus_details}.

The corpus is freely available \footnote{http://www.cs.man.ac.uk/\~{}filannim/timex3s\_corpus.csv} in CSV format using a tabulation character as delimiter.

\begin{table}
\centering
 \setlength{\tabcolsep}{17pt}
 \begin{tabular}{ll}
  \toprule
  \bf{Timex type} & {\bf Frequency} \\
  \midrule
  DATE & 2307 \\
  DURATION & 416 \\
  TIME & 71\\
  SET & 28\\
  \midrule
  TOTAL & 2822 \\
  \bottomrule
 \end{tabular}
\caption{Distribution of TIMEX3 tags in the corpus.}
\label{corpus_details}
\end{table}

\begin{table*}
\centering
 \setlength{\tabcolsep}{17pt}
 \begin{tabular}{llll}
  \toprule
  \bf{Temporal expression} & {\bf Type} & {\bf Value} & {\bf Utterance} \\
  \midrule
    \ldots & & & \\
    more than two years & DURATION & P2Y & 20110926 \\
    much of 2010 & DATE & FUTURE\_REF & 20110926 \\
    nearly a month & DATE & P1M & 20110926 \\
    nearly an hour & DURATION & PT1H & 19910225 \\
    nearly forty years & DURATION & P40Y & 1919980120 \\
    nearly four years ago & DATE & 1994 & 19980227:081300 \\
    nearly three years & DURATION & P3Y & 19891030 \\
    nearly two months & DURATION & P2M & 19980306:131900 \\
    nearly two months afterwards & DATE & FUTURE\_REF & 20110926 \\
    nearly two weeks ago & DATE & 1989-WXX & 19891030 \\
    nearly two years & DURATION & P2Y & 19980301:141100 \\
    next day & DATE & 2011-09-27 & 20110926 \\
    \ldots & & & \\
\end{tabular}
\caption{Brief excerpt of the corpus.}
\label{excerpt_corpus}
\end{table*}

\subsection{Temporal expressions normaliser}

I built a new normaliser on top of the one freely available from University of Rochester\footnote{http://www.cs.rochester.edu/u/naushad/temporal}: TRIOS. It is a rule-based normaliser and it has been proved to provide the second best performance in TempEval-2 Shared Task \cite{UzZaman10}. All the rules are in the form of regular expressions in a switch architecture: the activation of one of them excludes the activation of all the others.

I introduced a top layer with three new kinds of rules: extension, manipulation and post-manipulation rules.

The extension rules are just new rules that cover non-expected cases and are checked immediately before the pre-existing rules. If a temporal expression do not activate any of the extension rule, it goes into TRIOS. For example, some of these rules are used to normalise expressions of festivities dates such as \emph{``Thanksgiving day"} or \emph{``Saint Patrick's day"}.

The manipulation rules have been introduced to turn particular well-known expressions into an easier form before TRIOS processes them. Once one of these rules is activated, the original temporal expression is transformed into a reduced one that is easier to normalise properly for the pre-existing set of rules. After the transformation, the new temporal expression is taken in input by TRIOS for the normalisation task.

Lastly, I used the post-manipulation rules to solve some deficiencies in the normaliser by adding further information lost by TRIOS and finally improving the performance. In this case the temporal expression is evaluated through the extension rules or the original set. At the end of the normalisation process the result is enriched with further information. For example, I used these rules to add information about seasons which are not considered in TRIOS at all.

In the end, I introduced 32 new regular expression patterns: 16 extension rules, 12 manipulation rules and 4 post-manipulation rules. The entire system is freely available online\footnote{http://www.cs.man.ac.uk/\~{}filannim/timex\_normaliser.zip} under GNU licence\footnote{http://www.gnu.org/licenses/gpl.html}.

\section{Evaluation}

I evaluated the normalisation system using the new corpus previously described as a training set and then I measured the performances with respect to the TempEval-2 Shared Task test set. This offered me the possibility of comparing my normaliser with all the others evaluated in that challenge.

In order to measure the difference between TRIOS and my extension I also tested both of them by using the new corpus. It is important to notice that TRIOS has been trained on the same data provided in the new corpus. For this reason a comparison between these systems is legitimate.

In both cases, the evaluation procedure is based on counting. Because the normalisation task is aimed at providing the right \textsc{type} attribute and the right \textsc{value} attribute, the evaluation is carried out by counting how many times the system provides the same value with respect to the human ones. It is important to emphasise that every value provided by the system that differs form the human one for at least one character is considered error.

If this method is quite reasonable for \textsc{type} attribute, it might be too restrictive for \textsc{value} attribute. Some practical examples could be of help to explain the problem.
\begin{itemize}
\item The human annotation of a certain timex is \emph{\{type: "DATE", value: "FUTURE\_REF"\}} whereas the system provides a the more specific annotation \emph{\{type: "DATE", value: "2013-09-XX"\}}.
\item The system provides an annotation that is less specific than that provided by humans. For example, it happens when the human-annotation is \emph{\{type: "DATE", value: "2011-04-18"\}} and the system provide \emph{\{type: "DATE", value: "2011-04-XX"\}}. 
\end{itemize}
In all these cases the annotations are considered completely wrong. Even when the system provides a partially wrong annotation, e.g. \emph{\{type: "DATE", value: "2011-04-23"\}} for a human annotation of \emph{\{type: "DATE", value: "2011-04-18"\}}, considering it a complete wrong result may be too strict because year and month are correct however. This fact has justified the investigation of other measurement metrics \cite{UzZaman11}.
 
\subsection{Results}

The normalisation results with respect to TempEval-2 Shared Task are shown in Table \ref{tempeval2-comparison}. The new TRIOS extension outperforms each system in the normalisation of \textsc{value} attributes and performs competitively in the normalisation of \textsc{type} attributes.

The table already shows that the normalisation of value attributes is slightly harder than that of type attributes. The extension of TRIOS outperformed the original system of 2.81\% for \textsc{type} attribute and 9.13\% for \textsc{value} attribute.

I randomly sub-sampled (400 temporal expressions) the original corpus 10 times and I measured the performances with TRIOS and my extension. I conducted a statistical analysis on the results and I proved that the difference is statistically significant (Willcoxon test), respectively {\textrho} = 0.00586 and  {\textrho} = 0.0001621.

The normalisation results with respect to the new corpus are shown in Table \ref{new-corpus-comparisons}.

\begin{table}
\centering
 \setlength{\tabcolsep}{17pt}
 \begin{tabular}{llll}
  \toprule
  \bf{} & type & value \\
  \midrule
  Edinburgh & 0.84 & 0.63 \\
  HeidelTime & 0.96 & 0.85 \\
  KUL & 0.91 & 0.55 \\
  TERSEO & {\bf 0.98} & 0.65 \\
  TipSem & 0.92 & 0.65 \\
  TRIOS & 0.94 & 0.76 \\
  {\bf TRIOS extension} & 0.95 & {\bf 0.86} \\ 
  \bottomrule
 \end{tabular}
\caption{Results obtained from TempEval-2 test set.}
\label{tempeval2-comparison}
\end{table}

\begin{table}
\centering
 \setlength{\tabcolsep}{17pt}
 \begin{tabular}{llll}
  \toprule
  \bf{} & type & value \\
  \midrule
  TRIOS & 0.8572 & 0.6257 \\
  TRIOS extension & {\bf 0.8853} & {\bf 0.7170} \\
  \bottomrule
 \end{tabular}
\caption{Results obtained from the corpus.}
\label{new-corpus-comparisons}
\end{table}

\subsection{Error analysis}

\begin{table*}
\centering
 \setlength{\tabcolsep}{17pt}
 \begin{tabular}{lll}
  \toprule
  \bf{} & {\bf human} & {\bf system} \\
  \midrule
  25 & 1999-04-25 & n/a \\
  last year & 1988-Q2 & 1988 \\
  three years before & FUTURE\_REF & PAST\_REF \\
  the summer of 1862 & FUTURE\_REF & 1862-SU \\
  the weekend & P2D & PRESENT\_REF \\
  \bottomrule
 \end{tabular}
\caption{Some errors made by the normaliser.}
\label{errors}
\end{table*}

The original TRIOS normaliser made 1023 value mistakes and 402 type mistakes while its extension respectively made 779 and 323.
Through an accurate analysis of the errors, I found plenty of human annotations that seemed to be wrong at first impression. Once I analysed the same annotations taking into account the entire sentence from which each expression had been extracted, I found that the human annotations were actually right. Some examples are shown in Table \ref{errors}.

This leads to the conclusion that further improvements are possible only if I consider also the resolution of anaphoric expressions. To do this, it will be necessary to consider a wider window for each temporal expression that takes into account at least the entire sentence in which each temporal expression is located.

\section{Conclusions}

I introduced a new rule-based normaliser of temporal expressions and I showed that it resulted in better performances than the current state-of-the-art system with respect to TempEval-2 Shared Task. I also illustrated the corpus of temporal expressions for normalisation and its purpose. I made both, the normaliser and the corpus, freely available on-line (GNU public licence apply).

\subsection{Future work}

The work presented in this report is the product of a preliminary study in the field of information extraction. The results presented in this report clearly show the necessity of coping with anaphoric temporal expression to substantially enhance the performances of normalisation phase. Currently, the normalisation task takes into account only the temporal expressions, without considering a wider window, such as the entire sentence or a pre-defined number of words after and before the expression. This is required in order to cope with anaphoric expressions.

My long-term goal is to develop novel temporal expressions extraction techniques and use them in clinical domain. Because of the lack of pre-annotated clinical data, I will explore the use of semi-supervised machine learning approaches for the identification phase.

\subsection{Acknowledgements}

I would like to thank Naushad UzZaman from the University of Rochester to have shared his normaliser with the scientific community. I would also like to acknowledge the support of UK Engineering and Physical Science Research Council in the form of doctoral training grant.


\bibliography{bibliography}

\end{document}